\pdfoutput=1

\documentclass[11pt]{article}

\usepackage{EMNLP2023}

\usepackage{times}
\usepackage{latexsym}

\usepackage[T1]{fontenc}

\usepackage[utf8]{inputenc}

\usepackage{microtype}
\usepackage{inconsolata}

\usepackage{rotating}
\usepackage{blindtext}
\usepackage{amsmath}
\usepackage{amsfonts}
\usepackage{graphicx}
\usepackage{mathtools}
\usepackage{booktabs}
\usepackage{relsize}
\usepackage{float}
\usepackage{graphics}
\usepackage{multirow}
\usepackage{algorithm}
\usepackage{algpseudocode}
\usepackage[utf8]{inputenc}

\DeclareUnicodeCharacter{01B0}{\`u}
\DeclareUnicodeCharacter{01A1}{\`o}

\usepackage{threeparttable}
\usepackage{booktabs, caption, makecell}

\usepackage[title]{appendix}
\usepackage[cal=cm]{mathalfa}
\usepackage[colorinlistoftodos]{todonotes}
\usepackage{multirow}
\usepackage[export]{adjustbox}
\usepackage{subcaption}
\usepackage{pifont}
\newcommand{\ignore}[1]{}

\newcommand{\cmark}{\ding{51}}%
\newcommand{\xmark}{\ding{55}}%
\usepackage{microtype}
\usepackage[subtle]{savetrees}
\title{Transformers as Graph-to-Graph Models}

\author{James Henderson~$^{1}$ ~~Alireza Mohammadshahi~\thanks{~~Work done while working at Idiap Research Institute.}~~$^{1,2,3}$ ~~Andrei C. Coman~$^{1,2}$ \\ \textbf{Lesly Miculicich}~$^{*}$\thanks{~~Now at Google} \vspace{0.1cm}\\ 
 $^1$ Idiap Research Institute ~~~~~ $^2$ EPFL ~~~~~ $^3$ University of Zurich  \\
 \texttt{\{james.henderson,andrei.coman\}@idiap.ch}\\\texttt{alireza.mohammadshahi@epfl.com} \\\texttt{lmiculicich@google.com} 
}

\begin{document}
\maketitle

\begin{abstract}

We argue that Transformers are essentially graph-to-graph models, with sequences just being a special case.  Attention weights are functionally equivalent to graph edges.  Our Graph-to-Graph Transformer architecture makes this ability explicit, by inputting graph edges into the attention weight computations and predicting graph edges with attention-like functions, thereby integrating explicit graphs into the latent graphs learned by pretrained Transformers.  Adding iterative graph refinement provides a joint embedding of input, output, and latent graphs, allowing non-autoregressive graph prediction to optimise the complete graph without any bespoke pipeline or decoding strategy.
Empirical results show that this architecture achieves state-of-the-art accuracies for modelling a variety of linguistic structures, integrating very effectively with the latent linguistic representations learned by pretraining.

\end{abstract}

\section{Introduction}

Computational linguists have traditionally made extensive use of structured representations to capture the regularities found in natural language.  The huge success of Transformers~\cite{transformervaswani} and their pre-trained large language models~\cite{devlin-etal-2019-bert,zhang2022opt,touvron2023llama,touvron2023llama2} have brought these representations into question, since these models are able to capture even subtle generalisations about language and meaning in an end-to-end sequence-to-sequence model~\cite{wu-etal-2020-perturbed, michael-etal-2020-asking, hewitt-etal-2021-conditional}. This raises issues for research that still needs to model structured representations, such as work on knowledge graphs, hyperlink graphs, citation graphs, or social networks.

In this paper we show that the sequence-to-sequence nature of most Transformer models is only a superficial characteristic; underlyingly they are in fact modelling complex structured representations.  We survey versions of the Transformer architecture which integrate explicit structured representations with the latent structured representations of Transformers.  These models can jointly embed both the explicit structures and the latent structures in a Transformer's sequence-of-vectors hidden representation, and can predict explicit structures from this embedding.  In the process, we highlight evidence that the latent structures of pretrained Transformers 
already include much information about traditional linguistic structures.
These Transformer architectures support explicit structures which are general graphs, making them applicable to a wide range of structured representations and their integration with text.

The key insight of this line of work is that attention weights and graph structure edges are effectively the same thing.  Linguistic structures are fundamentally an expression of locality in the interaction between different components of a representation.  As \citet{henderson-2020-unstoppable} argued, incorporating this information about locality in the inductive bias of a neural network means putting connections between hidden vectors if their associated components are local in the structure.  In Transformers~\cite{transformervaswani}, these connections are learned in the form of attention weights.  Thus, these attention weights are effectively the induced structure of the Transformer's latent representation.

However, attention weights are not explicitly part of a Transformer's hidden representation.  The output of a Transformer encoder is a sequence of vectors, and the same is true of each lower layer of self-attention.  The latent attention weights are extracted from these sequence-of-vector embeddings with learned functions of pairs of vectors.  Edges in explicit graphs can be predicted in the same way (from pairs of vectors), assuming that these graphs have also been embedded in the sequence of vectors.

In recent years, the main innovation has been in how to embed explicit graphs in the hidden representations of Transformers.  In our work on this topic, we follow the above insight and input the edges of the graph into the computation of attention weights.  Attention weights are computed from an $n\times n$ matrix of attention scores~(where $n$ is the sequence length), so we input the label of the edge between nodes $i$ and $j$ into the score computation for the $i,j$ cell of this matrix.  Each edge label has a learned embedding vector, which is input to the attention score function in various ways depending on the architecture.  This allows the Transformer to integrate the explicit graph into its own latent attention graph in flexible and powerful ways.  This integrated attention graph can then determine the Transformer's sequence-of-vectors embedding in the same way as standard Transformers. 

Researchers from the Natural Language Understanding group at Idiap Research Institute have developed this architecture for inputting and predicting graphs under the name of \textit{Graph-to-Graph Transformer} (G2GT).  G2GT allows conditioning on an observed graph and predicting a target graph.  For the case where a graph is only observed at training time, we not only want to predict its edges, we also want to integrate the predicted graph into the Transformer embedding.  This has a number of advantages, most notably the ability to jointly model all the edges of the graph.
By iteratively refining the previous predicted graph, G2GT can jointly model the entire predicted graph even though the actual prediction is done independently for each edge.  And this joint modelling can be done in conjunction with other explicit graphs, as well as with the Transformer's induced latent graph.

Our work on G2GT has included a number of different explicit graph structures.  
The original methods were developed on syntactic parsing \cite{10.1162/tacl_a_00358,mohammadshahi-henderson-2020-graph,Mohammadshahi:303812}.  The range of architectures was further explored for semantic role labelling \cite{mohammadshahi-henderson-2023-syntax} and collocation recognition~\cite{espinosa-anke-etal-2022-multilingual}.  G2GT's application to coreference resolution extended the complexity of graphs to two levels of representation (mention spans and coreference chains) over an entire document, which was all modelled with iterative refinement of a single graph
\citep{miculicich-henderson-2022-graph}.  Current work on knowledge extraction poses further challenges, most notably the issue of tractably modelling large graphs.  The code for G2GT is open-source and available for other groups to use for other graph structures (at \url{https://github.com/idiap/g2g-transformer}).

In the rest of this paper, we start with a review of related work on deep learning for graph modelling (Section~\ref{sec:relatedwork}).  We then present the general G2GT architecture with iterative refinement (Section~\ref{sec:architecture}), before discussing the specific versions we have evaluated on specific tasks (Section~\ref{sec:models}).  We then discuss the broader implications of these results (Section~\ref{sec:discussion}), and conclude with a discussion of future work (Section~\ref{sec:conclusion}).

\section{Deep Learning for Graphs}
\label{sec:relatedwork}

\paragraph{Graph Neural Networks.} Early attempts at broadening the application of neural networks to graph structures were pursued by \citet{gori2005new} and \citet{scarselli2008graph}, who introduced the Graph Neural Networks (GNNs) architecture as a natural expansion of Recurrent Neural Networks (RNNs) \cite{hopfield1982neural}. This architecture regained interest in the context of deep learning, expanded through the inclusion of spectral convolution layers \cite{Bruna2013SpectralNA}, gated recurrent units \cite{Li2015GatedGS}, spatial convolution layers \cite{kipf2017semisupervised}, and attention layers \cite{velickovic2018graph}. GNNs generally employ the iterative local message passing mechanism to aggregate information from neighbouring nodes \cite{gilmer2017neural}. Recent research, analysing GNNs through the lens of \citet{weisfeiler1968reduction}, has highlighted two key issues: over-smoothing \cite{oono2020graph} and over-squashing \cite{alon2021on}. Over-smoothing arises from repeated aggregation across layers, leading to convergence of node features and loss of discriminative information. Over-squashing, on the other hand, results from activation functions during message aggregation, causing significant information and gradient loss. These issues limit the capacity of GNNs to effectively capture long-range dependencies and nuanced graph relationships \cite{Topping2021UnderstandingOA}.  
The Transformer architecture \cite{transformervaswani} can be seen as addressing these issues, in that 
its stacked layers of self-attention can be seen as a fixed sequence of learned aggregation steps.

\paragraph{Graph Transformers.} Transformers \cite{transformervaswani}, initially designed for sequence tasks, represent a viable and versatile alternative to GNNs due to their intrinsic graph processing capabilities. Through their self-attention mechanism, they can seamlessly capture global wide-ranging relationships, akin to handling a fully-connected graph. 
\citet{shaw-etal-2018-self} explicitly input relative position relations as embeddings into the attention function, thereby effectively inputting the relative position graph, instead of absolute position embeddings, to represent the sequence. Generalising this explicit input strategy to arbitrary graphs \citep{henderson-2020-unstoppable} has led to a general class of models which 
we will refer to as \textit{Graph Transformers} (GT).

\paragraph{GT Evolution and Applications.} The history of graph input methods used in GTs started with Transformer variations that experimented with relative positions to more effectively capture distance between input elements. 
Rather than adopting the sinusoidal position embedding introduced by \citet{transformervaswani} or the absolute position embedding proposed by \citet{devlin-etal-2019-bert}, \citet{shaw-etal-2018-self} added relative position embeddings to attention keys and values, capturing token distance within a defined range.  \citet{dai-etal-2019-transformer} proposed Transformer-XL, which used content-dependent positional scores and a global positional score in attention weights. \citet{mohammadshahi-henderson-2020-graph} demonstrated one of the earliest successful integration of an explicit graph into Transformer's latent attention graph. They introduced the \textit{Graph-To-Graph Transformer} (G2GT) architecture and applied it to syntactic parsing tasks by effectively leveraging pre-trained models such as BERT \cite{devlin-etal-2019-bert}. \citet{huang-etal-2020-improve} introduced new methods to enhance interaction between query, key and relative position embeddings within the self-attention mechanism. \citet{Su2021RoFormerET} proposed RoFormer, which utilises a rotation matrix to encode absolute positions while also integrating explicit relative position dependencies into the self-attention formulation. \citet{Liutkus2021RelativePE} and \citet{chen-2021-permuteformer} extended Performer \cite{Choromanski2020RethinkingAW} to support relative position encoding while scaling Transformers to longer sequences with a linear attention mechanism. Graphormer \cite{Ying2021DoTR} introduced node centrality encoding as an additional input level embedding vector, node distances and edges as soft biases added at attention level, and obtained excellent results on a broad range of graph representation learning tasks. \citet{10.1162/tacl_a_00358} built upon the G2GT architecture and proposed an iterative refinement procedure over previously predicted graphs, using a non-autoregressive approach. SSAN \cite{Xu2021EntitySW} leveraged the GT approach to effectively model mention dependencies for document-level relation extraction tasks. JointGT \cite{ke-etal-2021-jointgt} exploited the GT approach for knowledge to text generation tasks via a joint graph-text encoding. Similarly, TableFormer \cite{yang-etal-2022-tableformer} demonstrated the successful utilisation of the GT approach for combined text-table encoding in table-based question answering tasks. \citet{espinosa-anke-etal-2022-multilingual} proposed a GT architecture for simultaneous collocation extraction and lexical function typification, incorporating syntactic dependencies into the attention mechanism. \citet{miculicich-henderson-2022-graph} showed that the G2GT iterative refinement procedure can be effectively applied to graphs at multiple levels of representation. \citet{Diao2022RelationalAG} further extended a GT architecture with new edge and node update methods and applied them to graph-structured problems.
QAT \cite{Park2022RelationawareLT} substantially expanded upon GT models to jointly handle language and graph reasoning in question answering tasks. In the study conducted by \citet{mohammadshahi-henderson-2023-syntax}, the G2GT model showed substantial improvements in the semantic role labelling tasks.  
The multitude of successful applications and extensions firmly establish Graph Transformers as a robust and adaptable framework for addressing complex challenges in language and graphs.

\section{Graph-to-Graph Transformer Architecture}
\label{sec:architecture}

Our Graph-to-Graph Transformer (G2GT) architecture combines the idea of inputting graph edges into the self-attention function with the idea of predicting graph edges with an attention-like function.
By encoding the graph relations into the self-attention mechanism of Transformers, the model has an appropriate linguistic bias, without imposing hard restrictions. Specifically, G2GT modifies the attention mechanism of Transformers~\cite{transformervaswani} to input any graph. Given the input sequence~$W=(x_1,x_2,...,x_n)$, and graph relations~$G=\{(x_i,x_j,l),1 \leq i,j \leq n, l \in L\}$~(where $L$ is the set of labels), the modified self-attention mechanism is calculated as\footnote{Various alternative functions are possible for inputting relation embeddings into attention weight computations.  \citet{dufter-etal-2022-position} provide a survey of previous proposals for relative position encoding. In ongoing work, we have found that using a relation embedding vector to reweight the dimensions in standard dot-product attention works well for some applications.}:
\vspace{-3ex}
\begin{align}
\label{eq:g2g-attn}
\\[-1ex]
\begin{split}
e_{ij} = \frac{1}{\sqrt{d}} \Big[ x_i\boldsymbol{W^Q}(x_j\boldsymbol{W^K})^T
+x_i\boldsymbol{W^Q}(r_{ij}\boldsymbol{W^R_1})^T \\[-1ex]
+r_{ij}\boldsymbol{W^R_2}(x_j\boldsymbol{W^K})^T \Big]
\end{split}
\nonumber
\end{align}
where $r_{ij}\in \{0,1\}^{|L|}$ is a one-hot vector which specifies the type of the relation between $x_i$ and $x_j$,\footnote{This formulation can be easily extended to multi-label graphs by removing the one-hot constraint.  We are investigating the most effective method for doing this.} $\boldsymbol{W^R_1},\boldsymbol{W^R_2} \in R^{|L| \times d}$ are matrices of graph relation embeddings which are learned during training, $|L|$ is the label size, and $d$ is the size of hidden representations.
The value equation of Transformer~\cite{transformervaswani} is also modified to pass information about graph relations to the output of the attention function:
\vspace{-0.5ex}
\begin{align}
\label{eq:g2g-value}
z_i = \sum_j\alpha_{ij}(x_j\boldsymbol{W^V}+r_{ij}\boldsymbol{W^R_3})
\\[-4.5ex]\nonumber
\end{align}
where $\boldsymbol{W^R_3} \in R^{|L| \times d}$ is another learned relation embedding matrix.

To extract the explicit graph from the sequence of vectors output by the Transformer, a classification module is applied to pairs of vectors and maps them into the label space $L$. Initially, the module transforms each vector into distinct head and tail representations using dedicated projection matrices. Subsequently, a classifier (linear, bilinear or MLP) is applied, to map the vector pair onto predictions over the label space. Notably, each edge prediction can be computed in parallel (i.e.\ in a non-autoregressive manner), as predictions for each pair are independent of one another. Given the discrete nature of the output, various decoding methods can be employed to impose desired constraints on the complete output graph. These can range from straightforward head-tail order constraints, to more complex decoding algorithms such as the Minimum Spanning Tree (MST) algorithm.

Having an architecture which can both condition on graphs and predict graphs gives us the powerful ability to do iterative refinement of arbitrary graphs.  Even when graph prediction is non-autoregressive, conditioning on the previously predicted graph allows the model to capture between-edge correlations like an autoregressive model.
As illustrated in Figure~\ref{fig:rg2g_model}, we propose \textbf{R}ecursive \textbf{N}on-autoregressive \textbf{G}2G\textbf{T}~(RNGT), which predicts all edges of the graph in
parallel, and is therefore non-autoregressive, but 
can still condition every edge prediction on all other edge predictions by
conditioning on the previous version of the graph 
(using Equations~\ref{eq:g2g-attn} and \ref{eq:g2g-value}).

The input to the model is the input graph $W$ (e.g.\ a sequence of tokens), and the output is the final graph $G^T$ over the same set of nodes. First, we compute an initial graph $G^0$ over the nodes of $W$, which can be done with any model.  Then each recursive iteration encodes the previous graph $G^{t-1}$ and predicts a new graph $G^t$. It can be formalised in terms of an encoder $\operatorname{E^{RNG}}$ and a decoder $\operatorname{D^{RNG}}$:
\begin{figure}
  \centering
  \includegraphics[width=\linewidth]{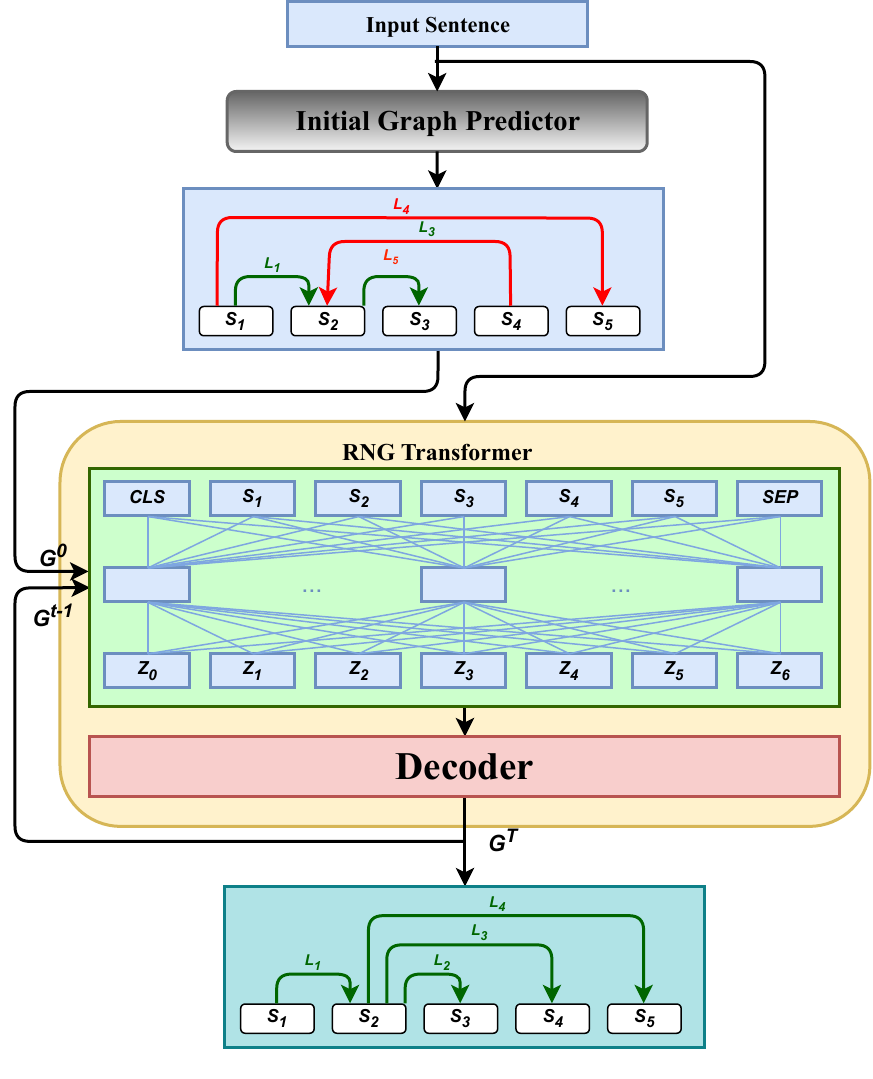}
  \vspace{-4ex}
  \caption{The Recursive Non-autoregressive Graph-to-Graph Transformer architecture.}
  \label{fig:rg2g_model}
\end{figure}
\begin{equation}
    \begin{cases}
        Z^t = \operatorname{ E^{RNG}}(W,G^{t-1}) \\
        G^t = \operatorname{ D^{RNG}}(Z^t) 
    \end{cases}
    t = 1,\ldots,T
\label{eq:rng-main}
\end{equation}
where $Z$ represents the set of vectors output by the model, and $T$ indicates the number of refinement iterations.  Note that in each step of this iterative refinement process, the G2G Transformer first computes a set of vectors which embeds the predicted graph (i.e.\ $\operatorname{ E^{RNG}}(W,G^{t-1})$), before extracting the edges of the predicted graph from this set-of-vectors embedding (i.e.\ $\operatorname{ D^{RNG}}(Z^t)$).

\section{G2GT Models and Results}
\label{sec:models}

This section provides a more comprehensive explanation of each alternative G2GT model we have explored, along with an outline of how we've applied these models to address various graph modelling problems.
The empirical success of these models demonstrate the computational adequacy of Transformers for extracting and modelling graph structures which are central to the nature of language.  The large further improvements gained by initialising with pretrained models demonstrates that Transformer pretraining encodes information about linguistic structures in its attention mechanisms.

\subsection{Syntactic Parsing}

Syntactic parsing is the process of analysing the grammatical structure of a sentence, including identifying the subject, verb, and object. Syntactic dependency parsing is a critical component in a variety of natural language understanding tasks, such as semantic role labelling~\citep{henderson-etal-2013-multilingual,marcheggiani-titov-2017-encoding,marcheggiani-titov-2020-graph}, machine translation~\citep{Chen_2017}, relation extraction~\citep{zhang-etal-2018-graph}, and natural language inference~\citep{pang2019improving}.  
It is also a benchmark structured prediction task, because architectures which are not powerful enough to learn syntactic parsing cannot be computationally adequate for language understanding. 

Syntactic structure is generally specified in one of two popular grammar styles, constituency parsing~(i.e.\ phrase-structure parsing)~\citep{manning1999foundations, henderson-2003-inducing, henderson-2004-discriminative, titov-henderson-2007-constituent} and dependency parsing~\citep{nivre-2003-efficient, titov-henderson-2007-latent, carreras-2007-experiments,nivre-mcdonald-2008-integrating,dyer-etal-2015-transition,msh2021}. There are two main approaches to compute the dependency tree: transition-based and graph-based parsers. Transition-based parsers predict the dependency graph one edge at a time through a sequence of parsing actions \citep{yamada-matsumoto-2003-statistical,nivre-scholz-2004-deterministic,titov-henderson-2007-latent,zhang-nivre-2011-transition,weiss-etal-2015-structured,yazdani-henderson-2015-incremental}, and graph-based parsers compute scores for every possible dependency edge and then apply a decoding algorithm to find the highest scoring total tree \citep{mcdonald-etal-2005-online,koo2010efficient,kuncoro-etal-2016-distilling,zhou-zhao-2019-head}. 

In the following, we outline our proposals for using G2GT for syntactic parsing tasks.

\subsubsection{Transition-based Dependency Parsing}

In \citep{mohammadshahi-henderson-2020-graph}, we integrate the G2GT model with two baselines, named StateTransformer (StateTr) and SentenceTransformer (SentTr). In the former model, we directly input the parser state into the G2GT model, while the latter takes the initial sentence as the input. For better efficiency of our transition-based model, we used an alternative version of G2GT, introduced in Section~\ref{sec:architecture}, where the interaction of graph relations with key matrices in Equation~\ref{eq:g2g-attn} is removed.  Each parser decision is conditioned on the history of previous decisions by inputting an unlabelled partially constructed dependency graph to the G2GT model.  \citet{mohammadshahi-henderson-2020-graph} evaluate the integrated models on the English Penn Treebank~\cite{marcus-etal-1993-building}, and 13 languages of Universal Dependencies Treebanks~\cite{11234/1-2895}.

\begin{table}[t]
    \footnotesize
	\centering
	\tabcolsep=0.1cm
	\begin{tabular}{|lcc|}\hline
		 Model & ~UAS & ~LAS \\ \hline
        \newcite{andor-etal-2016-globally} & 94.61 & 92.79 \\
		\hline
		StateTr & 92.32 & 89.69 \\
		StateTr+G2GT & 93.07 & 91.08 \\[0.5ex]
		BERT StateTr & 95.18 & 92.73 \\
		BERT StateTr+G2GT & \textbf{95.58} & \textbf{93.74}\\
        \hline
		BERT SentTr & 95.65 & 93.85 \\
		BERT SentTr+G2GT & \textbf{96.06} & \textbf{94.26} \\
		\hline
	    \end{tabular}
        \vspace{-1.5ex}
	\caption{\label{tab:wsj-prev} 
	  Comparisons to the previous comparable models, including transition-based and sequence-to-sequence approaches~(according to \citet{mohammadshahi-henderson-2020-graph}) on English WSJ Treebank Stanford dependencies. Labelled and Unlabelled Attachment Scores~(LAS,UAS) are used as evaluation metrics.
        }
\end{table}

\ignore{
\begin{table}[t]
	\centering
	\begin{tabular}{|@{~~}l@{~}|@{~~}c@{~}|@{~~}c@{~}|@{~~}c@{~}|}
		\hline
		\!Language & \parbox{\widthof{Kulmizev et al.}}{\addtolength{\baselineskip}{-0.5ex}~\\ \newcite{kulmizev-etal-2019-deep}} & \parbox{\widthof{StateTr+G2GT}}{\addtolength{\baselineskip}{-0.5ex}~\\ BERT StateTr+G2GT} & \parbox{\widthof{SentTr+G2GT}}{\addtolength{\baselineskip}{-0.5ex}~\\ BERT SentTr+G2GT}\\[1.5ex]
        \hline
         \vspace{-2ex}
         & & & \\
		Arabic & {81.9} & {82.63} & {\textbf{83.65}} \\
		Basque & {77.9} & {74.03} & {\textbf{83.88}} \\ 
		Chinese & {83.7} & {85.91} & {\textbf{87.49}} \\
		English & {87.8} & {89.21} & {\textbf{90.35}} \\
		Finnish & {85.1} & {80.87} & {\textbf{89.47}} \\
		Hebrew & {85.5} & {87.0} & {\textbf{88.75}} \\
		Hindi & {89.5} & {93.13} & {\textbf{93.12}} \\
		Italian & {92.0} & {92.6} & {\textbf{93.99}} \\
		Japanese & {92.9} & {95.25} & {\textbf{95.51}} \\
		Korean & {83.7} & {80.13} & {\textbf{87.09}} \\
		Russian & {91.5} & {92.34} & {\textbf{93.30}} \\
		Swedish & {87.6} & {88.36} & {\textbf{90.40}} \\
		Turkish & {64.2} & {56.87} & {\textbf{67.77}} \\
		\hline
		Average & {84.87} & {84.48} & {\textbf{88.06}} \\
		\hline
	\end{tabular}
	\caption{\label{tab:ud-best} Labelled attachment score on 13 UD corpora for \newcite{kulmizev-etal-2019-deep} with BERT pre-training, BERT StateTr+G2GT, and BERT SentTr+G2GT models.
          \vspace{-1ex}
        } 
\end{table}
}

Results of our models on the Penn Treebank are shown in Table~\ref{tab:wsj-prev}
(see \citep{mohammadshahi-henderson-2020-graph} for further results on UD Treebanks).
Integrating the G2GT model with the StateTr baseline achieves 9.97\% LAS Relative Error Reduction~(RER) improvement, which confirms the effectiveness of modelling the graph information in the attention mechanism. Furthermore, initialising our model weights with the BERT model~\cite{devlin-etal-2019-bert}, provides significant improvement~(27.65\% LAS RER), which shows the compatibility of our modified attention mechanism with the latent representations learned by BERT pretraining. Integrating the G2GT model with the SentTr baseline results in a similar significant improvement~(4.62\% LAS RER). 

\subsubsection{Graph-based Dependency Parsing}

The StateTr and SentTr models generate the dependency graph in an autoregressive manner, predicting each parser action conditioned on the history of parser actions.  Many previous models have achieved better results with graph-based parsing methods, which use non-autoregressive computation of scores for all individual candidate dependency relations and then use a decoding method to reach the maximum scoring structure~\citep{mcdonald-etal-2005-online,koo2010efficient,ballesteros-etal-2016-training,wang-chang-2016-graph,kuncoro-etal-2016-distilling,zhou-zhao-2019-head}. However, these models usually ignore correlations between edges while predicting the complete graph. In \citep{10.1162/tacl_a_00358}, we propose the \textbf{R}ecursive \textbf{N}on-autoregressive \textbf{G}raph-to-Graph \textbf{Tr}ansformer (RNGT) architecture, as discussed in Section~\ref{sec:architecture}. The RNGT architecture can be applied to any task with a sequence or graph as input and a graph over the same set of nodes as output. Here, we apply it for the syntactic dependency parsing task, and preliminary experiments showed that removing the interaction of graph relations with key vectors, in Equation~\ref{eq:g2g-attn}, results in better performance and a more efficient attention mechanism.  \citet{10.1162/tacl_a_00358} evaluate this RNGT model on Universal Dependency (UD) Treebanks~\cite{11234/1-2895}, Penn Treebanks~\cite{marcus-etal-1993-building}, and the German CoNLL 2009 Treebank~\cite{hajic-etal-2009-conll} for the syntactic dependency parsing task.

\begin{table*}[bt]
  \centering\small
  \begin{tabular}{|c|cl|cl|c|}
    \hline
    \multirow{2}{*}{Language} & Multi & Multi+Mono & Mono & Mono & Mono \\
    & UDify & UDify+RNGT & SynTr & SynTr+RNGT & Empty+RNGT \\
    \hline
    Arabic & 82.88 & \textbf{85.93} (+17.81\%) & \textbf{86.23} & \textbf{86.31} (+0.58\%) & \textbf{86.05} \\
    Basque & 80.97 & 87.55 (+34.57\%) & 87.49 & \textbf{88.2} (+5.68\%) & \textbf{87.96} \\
    Chinese & 83.75 & 89.05 (+32.62\%) & 89.53 & \textbf{90.48} (+9.08\%)  & 89.82 \\
    English & 88.5 & 91.23 (+23.74\%) & \textbf{91.41} & \textbf{91.52} (+1.28\%) & 91.23 \\
    Finnish & 82.03 & \textbf{91.87} (+54.76\%) & \textbf{91.80} & \textbf{91.92} (+1.46\%) & \textbf{91.78} \\
    Hebrew & 88.11 & 90.80 (+22.62\%) & \textbf{91.07} & \textbf{91.32} (+2.79\%) & 90.56 \\
    Hindi & 91.46 & 93.94 (+29.04\%) & 93.95 & \textbf{94.21} (+4.3\%) & 93.97 \\
    Italian & 93.69 & 94.65 (+15.21\%) & \textbf{95.08} & \textbf{95.16} (+1.62\%) & \textbf{94.96} \\
    Japanese & 92.08 & \textbf{95.41} (+42.06\%) & \textbf{95.66} & \textbf{95.71} (+1.16\%) & \textbf{95.56} \\
    Korean & 74.26 & \textbf{89.12} (+57.73\%) & \textbf{89.29} & \textbf{89.45} (+1.5\%) & \textbf{89.1} \\
    Russian & 93.13 & \textbf{94.51} (+20.09\%) & \textbf{94.60} & \textbf{94.47} (-2.4\%) & 94.31 \\
    Swedish & 89.03 & 92.02 (+27.26\%) & 92.03 & \textbf{92.46} (+5.4\%) & \textbf{92.40} \\
    Turkish & 67.44 & 72.07 (+14.22\%) & \textbf{72.52} & \textbf{73.08} (+2.04\%) & 71.99 \\
    \hline
    Average & 85.18 & 89.86 & 90.05 & 90.33 & 89.98 \\
    \hline
  \end{tabular}
  \caption{\label{ud-las} Labelled attachment scores of monolingual~(SynTr) and multilingual~(UDify~\cite{kondratyuk-straka-2019-75}) baselines, and the refined models (+RNGT) pre-trained with BERT~\cite{devlin-etal-2019-bert} on 13 languages of UD Treebanks. The relative error reduction after the integration is illustrated in parentheses. Bold scores are not significantly different from the best score in that row (with $\alpha=0.01$).
  }
\end{table*}

Table~\ref{ud-las}
shows the results on 13 languages of UD Treebanks. First, we use UDify~\cite{kondratyuk-straka-2019-75}, the previous state-of-the-art multilingual dependency parser,  as the initial parser for the RNGT model. The integrated model achieves significantly better LAS performance than the UDify model in all languages, which demonstrates the effectiveness of the RNGT model at refining a dependency graph. Then, we combine RNGT with Syntactic Transformer~(SynTr), a stronger monolingual dependency parser, which has the same architecture as the RNGT model except without the graph input mechanism. The SynTr+RNGT model reaches further improvement over the strong SynTr baseline~(four languages are significant), which is stronger evidence for the effectiveness of the graph refinement method. Interestingly, there is little difference between the performance with different initial parsers, implying that the RNGT model is effective enough to refine any initial graphs.  In fact, even when we initialise with an empty parse, the Empty+RNGT model achieves competitive results with the other RNGT models, again confirming our powerful method of graph refinement.

\ignore{
\subsubsection{Penn Treebank and German corpus Results}

UAS and LAS results for the Penn Treebanks and German CoNLL 2009 Treebank are reported in Table~\ref{penn-results}.
We compare to the results of previous state-of-the-art models and SynTr, and we use the RNGT model to refine both the Biaffine parser~\cite{dozat2016deep} and SynTr, on all Treebanks.\footnote{Results are calculated with the official evaluation script: (\url{https://depparse.uvt.nl/}). For German, we use \url{https://ufal.mff.cuni.cz/conll2009-st/eval-data.html}.}

Again, the SynTr model significantly outperforms previous state-of-the-art models, with a 5.78\%, 9.15\%, and 23.7\% LAS relative error reduction in English, Chinese, and German, respectively.  Despite this level of accuracy, adding RNGT refinement improves accuracy further under both UAS and LAS.  For the Chinese Treebank, this improvement is significant, with a 5.46\% LAS relative error reduction.
When RNGT refinement is applied to the output of the Biaffine parser~\cite{dozat2016deep}, it achieves a LAS relative error reduction of 10.64\% for the English Treebank, 16.05\% for the Chinese Treebank, and 27.72\% for the German Treebank.
These improvements, even over such strong initial parsers, again demonstrate the effectiveness of the RNGT architecture for graph refinement.
}

\subsection{Semantic Role Labelling}

The semantic role labelling (SRL) task provides a shallow semantic representation of a sentence and builds event properties and relations among relevant words, and is defined in both dependency-based~\cite{surdeanu-etal-2008-conll} and span-based~\cite{carreras-marquez-2005-introduction, pradhan-etal-2012-conll} styles. Previous work~\cite{marcheggiani-titov-2017-encoding,strubell-etal-2018-linguistically,cai-lapata-2019-semi,Fei_Li_Li_Ji_2021,zhou-etal-2020-parsing} showed that the syntactic graph helps SRL models to predict better output graphs, but finding the most effective way to incorporate the auxiliary syntactic information into SRL models was still an open question. In \citep{mohammadshahi-henderson-2023-syntax}, we introduce the Syntax-aware Graph-to-Graph Transformer
(SynG2G-Tr) architecture. The model conditions
on the sentence’s dependency structure and jointly
predicts both span-based~\cite{carreras-marquez-2005-introduction} and dependency-based~\cite{hajic-etal-2009-conll}
SRL structures. Regarding the self-attention mechanism, we remove the interaction of graph embeddings with value vectors in Equation~\ref{eq:g2g-value}, as it reaches better performance in this particular task \citep{mohammadshahi-henderson-2023-syntax}.

\begin{table}[t]
  \small
  \begin{tabular}{|lcc|} \hline
  Model & in-domain & out-of-domain \\
    \hline
    \textbf{end-to-end} & & \\
    \newcite{strubell-etal-2018-linguistically} & 84.99 & 74.66 \\
    SynG2G-Tr (w/o BERT) & \textbf{85.45} & \textbf{75.26} \\
    \hline
    \textit{+pre-training} & & \\
    \newcite{strubell-etal-2018-linguistically} & 86.9 & 78.25 \\
    SynG2G-Tr & \textbf{87.57} & \textbf{80.53} \\
    \hline
    \textbf{given predicate} & & \\
    \newcite{strubell-etal-2018-linguistically} & 86.04 & 76.54 \\
    SynG2G-Tr (w/o BERT) & \textbf{86.50} & \textbf{77.45} \\
    \hline
    \textit{+pre-training} & & \\
    \newcite{Jia_Yan_Wu_Tu_2022} & 88.25 & 81.90 \\
    SynG2G-Tr & \textbf{88.93} & \textbf{83.21} \\
    \hline
  \end{tabular}
  \caption{Comparing our SynG2G-Tr with previous comparable SoTA model on CoNLL 2005 test sets for both in-domain~(WSJ), and out-of-domain~(Brown) sets. Scores being boldfaced means that they are significantly better.\label{tab:srl}}
      \vspace{-1ex}
\end{table}

Results for span-based SRL are shown in Table~\ref{tab:srl}.
Without initialising the models with BERT~\cite{devlin-etal-2019-bert}, the SynG2G-Tr model outperforms a previous comparable state-of-the-art model~\cite{strubell-etal-2018-linguistically} in both \textit{end-to-end} and \textit{given-predicate} scenarios. The improvement indicates the benefit of encoding the graph information in the self-attention mechanism of Transformer with a soft bias, instead of hard-coding the graph structure into deep learning models~\cite{marcheggiani-titov-2017-encoding,strubell-etal-2018-linguistically,Xia_Li_Zhang_Zhang_Fu_Wang_Si_2019}, as the model can still learn other attention patterns in combination with this graph knowledge.  BERT~\cite{devlin-etal-2019-bert} initialisation results in further significant improvement in both settings, which again shows the compatibility of the G2GT modified self-attention mechanism with the latent structures learned by BERT pretraining.

\subsection{Coreference Resolution}

Coreference resolution (CR) is an important and complex task which is necessary for higher-level semantic representations.  We show that it benefits from a graph-based global optimisation of all the coreference chains in a document.

\subsubsection{CR Task Definition and Background}

Coreference resolution is the task of linking all linguistic expressions in a text that refer to the same entity. 
Solutions for this task involve three parts: mention-detection \cite{yu-etal-2020-neural, miculicich-henderson-2020-partially}, classification or ranking of mentions, and finally reconciling the decisions to create entity chains. 
These approaches fall within three principal categories: mention-pair models which perform binary decisions \cite{McCarthy1995UsingDT, aone-william-1995-evaluating, soon-etal-2001-machine}, entity-based models which focus on maintaining single underlying entity representation, contrasting the independent pair-wise decisions of mention-pair approaches \cite{clark-manning-2015-entity, clark-manning-2016-improving}, and ranking models which aim at ranking the possible antecedents of each mention instead of making binary decisions \cite{wiseman-etal-2016-learning}. A limitation of these methods lies in their bottom-up construction, resulting in an underutilisation of comprehensive global information regarding coreference links among all mentions in individual decisions. Furthermore, these methods tend to exhibit significant complexity. Modelling of coreference resolution as a graph-based approach offer an alternative to deal with these limitations. 

\begin{figure}[tb]
	\center
	\includegraphics[width=1\linewidth]{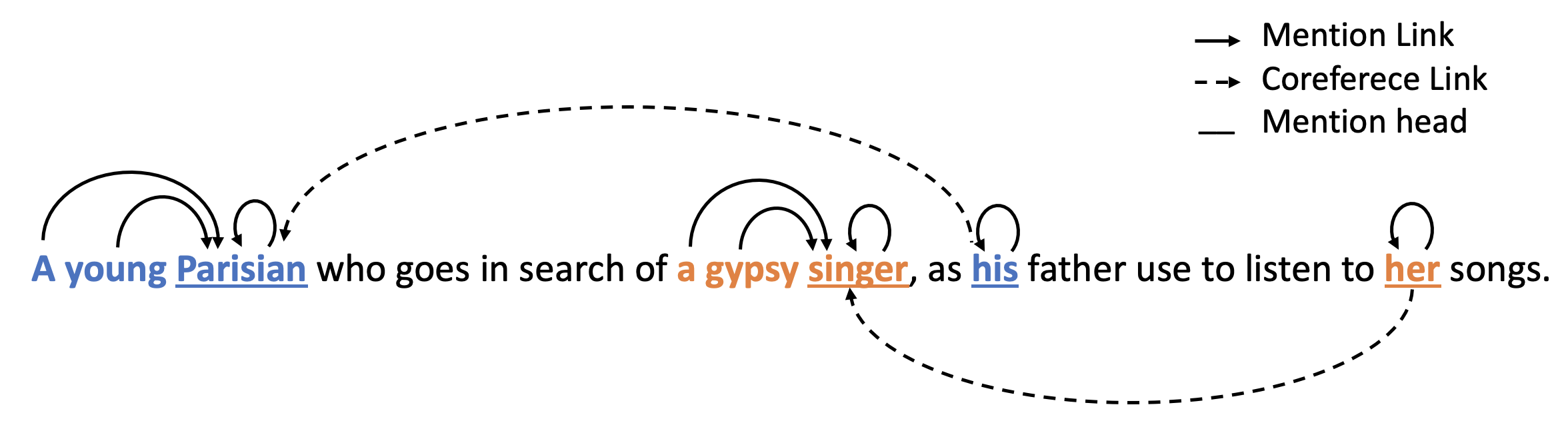}
        \vspace{-4ex}
	\caption{Example of a graph structure for coreference. Mention spans are shown in bold, and colours represent entity clusters. The mention heads are underlined.}
	\label{fig:coreference}
\end{figure}

\begin{figure}[tb]
	\center
	\includegraphics[width=1\linewidth]{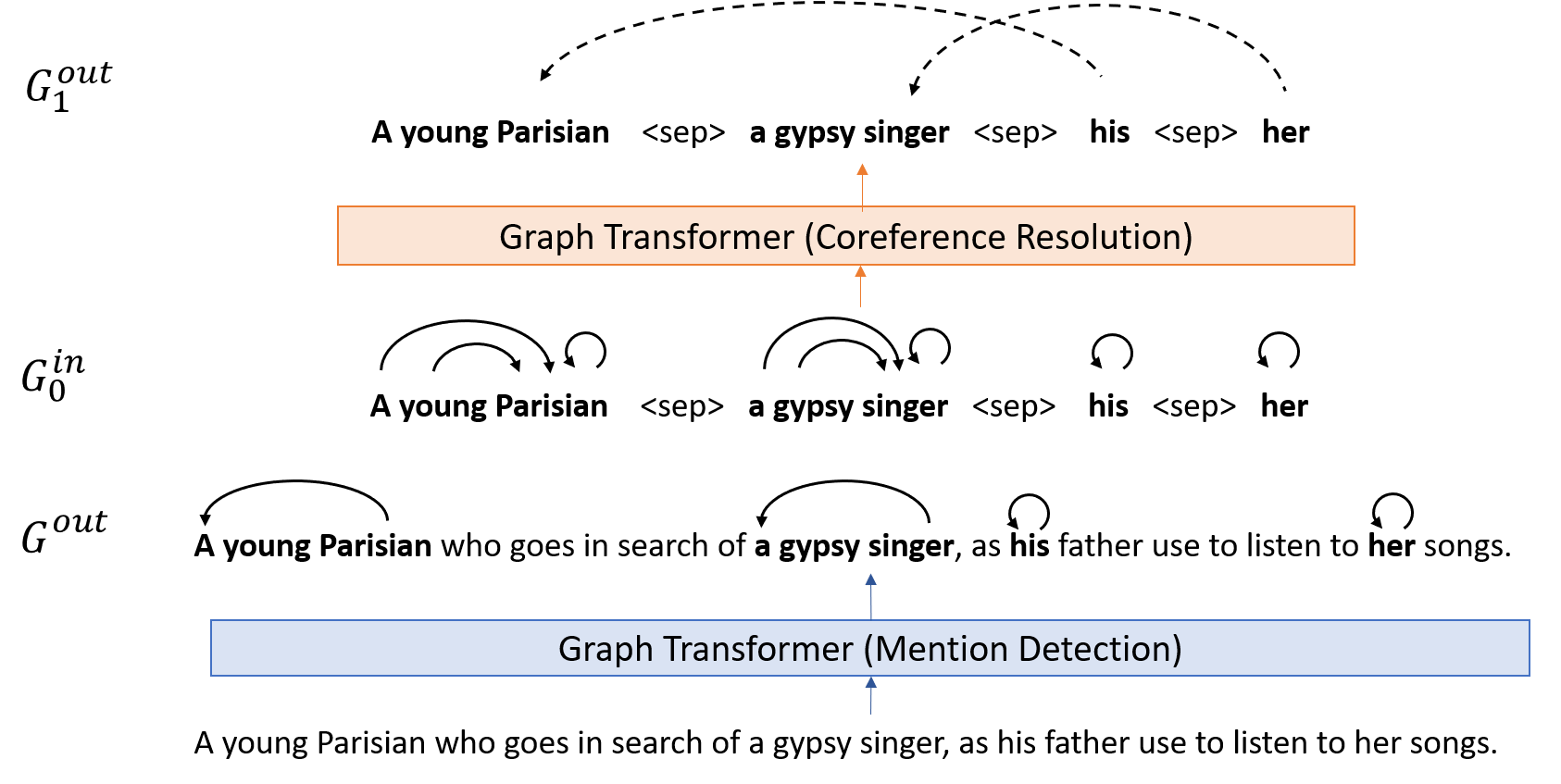}
	\caption{Example of iterations with G2GT in two stages.}
	\label{fig:coreference2}
\end{figure}

\subsubsection{Iterative Graph-based CR}

\citet{miculicich-henderson-2022-graph} proposed a novel approach to modelling coreference resolution, treating it as a graph problem. In this framework, the tokens within the text serve as nodes, and the connections between them signify coreference links (see Figure~\ref{fig:coreference}). Given a document $D=[x_1,...,x_N]$ with length $N$, the coreference graph is formally defined as the matrix $G \subset \mathbb{N}^{N \times N}$, which represents the relationships between tokens. Specifically, the relationship type between any two tokens, $x_i$ and $x_j$, is 
labelled as $g_{i,j} \in \{0,1,2\}$ for the three distinct relation types: (0) no link, (1) mention link, and (2) coreference link.

\begin{table*}[bt]
	\centering\small
	\begin{adjustbox}{width=1\textwidth}
		\begin{tabular}{@{~} l@{~~} c c c | c c c | c c c | c @{~}} 
			\toprule \
			& \multicolumn{3}{c}{\textbf{MUC}} & \multicolumn{3}{c}{$\mathbf{B^3}$} & \multicolumn{3}{c}{\textbf{CEAF$_{\phi_4}$}} \\ 
			\textbf{Model} & \textbf{P} & \textbf{R} & \textbf{F1} & \textbf{P} & \textbf{R} & \textbf{F1} & \textbf{P} & \textbf{R} & \textbf{F1} &\textbf{Avg. F1} \\ \hline
			\citet{lee-etal-2017-end} & 78.4 & 73.4 & 75.8 & 68.6 & 61.8 & 65.0 & 62.7 & 59.0 & 60.8 &  67.2  \\ 
			\citet{xu-choi-2020-revealing} & \textbf{85.9} & 85.5 &85.7 &79.0 &78.9 &79.0 &\textbf{76.7}& 75.2& 75.9& 80.2 \\
			\hline
			Baseline \cite{lee-etal-2018-higher}   & 81.4 & 79.5 & 80.4 & 72.2 & 69.5 & 70.8 & 68.2 & 67.1 & 67.6 & 73.0 \\
			+ BERT-large \cite{joshi-etal-2019-bert}  & 84.7 & 82.4 & 83.5 & 76.5 & 74.0 & 75.3 & 74.1 & 69.8 & 71.9 & 76.9 \\  
			+ SpanBERT-large \cite{joshi-etal-2020-spanbert} & 85.8 & 84.8 &85.3 &78.3 &77.9 &78.1& 76.4& 74.2 &75.3 &79.6 \\ \hline
			G2GT BERT-large \emph{reduced} & 84.7 & 83.1 & 83.9 & 76.8 & 74.0 & 75.4 & 75.3 & 70.1 & 72.6 & 77.3 \\
			G2GT SpanBERT-large \emph{reduced} & \textbf{85.9}	&\textbf{86.0}$^{*\dagger}$&	\textbf{85.9}$^{*}$&	\textbf{79.3}$^{*}$&	\textbf{79.4}$^{*\dagger}$&	\textbf{79.3}$^{*}$&	76.4&	\textbf{75.9}$^{*}$&	\textbf{76.1}$^{*}$&	\textbf{80.5}$^{*}$ \\ 
			\bottomrule 
		\end{tabular} 
	\end{adjustbox}  
	\caption{Evaluation of CR on the test set (CoNLL 2012) in terms of precision (P), recall (R) and F1 score for three metrics, as well as the average F1 over metrics.~  $*$ significant at p < 0.01 compared to \citep{joshi-etal-2020-spanbert},~ $\dagger$ significant at p < 0.05 compared to \citep{xu-choi-2020-revealing}.}
	\label{tab:coreference:results}
	
\end{table*}

The primary objective of this approach is to learn the conditional probability distribution $p(G|D)$. To achieve this, an iterative refinement strategy is employed, which captures interdependencies among relations. The model iterates over the same document $D$ for a total of $T$ iterations. In each iteration $t$, the predicted coreference graph $G_t$ is conditioned on the previous prediction, denoted as $G_{t-1}$. Thus, the conditional probability distribution of the model is defined as follows:
\begin{equation}
p(G^t|D, G^{t-1}) = \prod_{i=1}^N \prod_{j=1}^i p(g_{i,j}|D, G^{t-1})
\end{equation}

The proposed model operates on two levels of representation. In each iteration, it predicts the entire graph. However, during the first iteration, the model focuses on predicting edges that pinpoint mention spans, given that coreferent links only have relevance when mentions are detected. From the second iteration, both mention links, and coreference links are refined. This iterative strategy permits the model to enhance mention-related decisions based on coreference resolutions, and vice versa. This framework utilises iterative graph refinement as a substitute for conventional pipeline architectures in multi-level deep learning models. The iterative process concludes either when the graph no longer undergoes changes or when a predetermined maximum iteration count is attained (see Figure~\ref{fig:coreference2}).

Ideally, encoding the entirety of the document in a single pass would be optimal. However, in practical scenarios, a constraint on maximum length arises due to limitations in hardware memory capacity. To address this challenge, \citet{miculicich-henderson-2022-graph} introduce two strategies: overlapping windows and reduced document approach. In the latter strategy, mentions are identified during an initial iteration with a focus on optimising recall, as previously suggested in \cite{miculicich-henderson-2020-partially}. Only the representations of these identified spans are subsequently used as inputs for the following iterations.

\citet{miculicich-henderson-2022-graph} conducted experiments on the CoNLL 2012 corpus \citep{pradhan-etal-2012-conll} and showed improvements over relevant baselines and previous state-of-the-art methods, summarised in Table~\ref{tab:coreference:results}.  We compare our model with three baselines: \citet{lee-etal-2017-end} proposed the first end-to-end model for coreference resolution; \citet{lee-etal-2018-higher} extended the previous model by introducing higher order inference; and \citet{xu-choi-2020-revealing} used the span based pre-trained model SpanBERT \citep{joshi-etal-2020-spanbert}.  The `Baseline' of \citet{lee-etal-2018-higher} uses ELMo \cite{peters-etal-2018-deep} to obtain token representations, so versions of this Baseline which use `BERT-large' \citep{joshi-etal-2019-bert} and `SpanBERT-large' \citep{joshi-etal-2020-spanbert} as their pretrained models, are directly comparable to our `G2GT BERT-large' and `G2GT SpanBERT-large' models, respectively.

These results show that coreference resolution benefits from making global coreference decisions using document-level information, as supported by the G2GT architecture.
Our model achieves its optimal solution within a maximum of three iterations. Notably, due to the model's ability to predict the entire graph in a single iteration, its computational complexity is lower compared to that of the baseline approaches.

\section{Discussion}
\label{sec:discussion}

The empirical success of Graph-to-Graph Transformers on modelling these various graph structures helps us understand how Transformers model language.  This success demonstrates that Transformers are computationally adequate for modelling linguistic structures, which are central to the nature of language.  The reliance of these G2GT models on using self-attention mechanisms to extract and encode these graph relations shows that self-attention is crucial to how Transformers can do this modelling. 
The large improvements gained by initialising with pretrained models indicates that pretrained Transformers are in fact using the same mechanisms to learn about this linguistic structure, but in an unsupervised fashion.  

These insights into pretrained Transformers give us a better understanding of the current generation of Large Language Models (LLMs).  It is not that these models do not need linguistic structure (since their attention mechanisms do learn it); it is that these models do not need supervised learning of linguistic structure.  But perhaps in a low-resource scenario LLMs would benefit from the inductive bias provided by supervised learning of linguistic structures, such as for many of the world's languages other than English.  And these insights are potentially relevant to the issues of interpretability and controllability of LLMs.

These insights are also relevant for any applications which could benefit from integrating text with structured representations.  Our current work investigates jointly embedding text and parts of a knowledge base in a single G2GT model, providing a way to integrate interpretable structured knowledge with knowledge in text.  Such representations would be useful for information extraction, question answering and information retrieval, amongst many other applications.  Other graphs we might want to model with a Transformer and integrate with text include hyperlink graphs, citation graphs, and social networks.  An important open problem with such models is the scale of the resulting Transformer embedding.

\section{Conclusion and Future Work}
\label{sec:conclusion}

The Graph-to-Graph Transformer architecture makes explicit the implicit graph processing abilities of Transformers, but
further research is needed to fully leverage the potential of G2GT.

\subsection{Conclusions}

The success of the above models of a variety linguistic structures shows that Transformers are underlyingly graph-to-graph models, not limited to sequence-to-sequence tasks.  The G2GT architecture with its RNGT method provides an effective way to exploit this underlying ability when modelling explicit graphs, effectively integrating them with the implicit graphs learned by pre-trained Transformers. Inputting graph relations as features to the self-attention mechanism enables the information input to the model to be steered by domain-specific knowledge or desired outcomes but still learned by the Transformer, opening up the possibility for a more tailored and customised encoding process.
Predicting graph relations with attention-like functions and then re-inputting them for iterative refinement, encodes the input, predicted and latent graphs in a single joint Transformer embedding which is effective for making global decisions about structure in a text.

\subsection{Future Work}

One topic of research where explicit graphs are indispensable is knowledge graphs. Knowledge needs to be interpretable, so that it can be audited, edited, and learned by people.   And it needs to be integrated with existing knowledge graphs.  Our current work uses G2GT to integrate knowledge graphs with knowledge conveyed by text.


One of the limitations of the models discussed in this paper is that the set of nodes in the output graph needs to be (a subset of) the nodes in the input graph.  General purpose graph-to-graph mappings would require also predicting a set of new nodes in the output graph.  One natural solution would be autoregressive prediction of one node at a time, as is done for text generation, but an exciting alternative would be to use methods from non-autoregressive text generation in combination with our iterative refinement method RNGT.  

The excellent performance of the models presented in this paper suggest that many more problems can be successfully formulated as graph-to-graph problems and modelled with G2GT, in NLP and beyond.
The code for G2GT and RNGT is open-source and publicly available at \url{https://github.com/idiap/g2g-transformer}.

\section*{Acknowledgement}

We would like to especially thank the Swiss National Science Foundation for funding this work, under grants 200021E\_189458, CRSII5\_180320, and 200021\_178862.  We would also like to thank other members of the the Natural Language Understanding group at Idiap Research Institute for useful discussion and feedback, including Florian Mai, Rabeeh Karimi, Andreas Marfurt, Melika Behjati, and Fabio Fehr.


\bibliographystyle{acl_natbib}
\bibliography{emnlp2023,anthology}


\end{document}